\documentclass{article}
\usepackage[utf8]{inputenc}
\usepackage{graphicx}
\usepackage{subcaption}
\usepackage{url}
\usepackage{hyperref}
\usepackage{authblk}
\usepackage{xcolor}  
\usepackage[export]{adjustbox}

\title{Datasets for Studying Generalization from Easy to Hard Examples}

\author[1]{Avi Schwarzschild\footnote{Correspondence to \texttt{avi1@umd.edu}.}}
\author[2]{Eitan Borgnia}
\author[3]{Arjun Gupta}
\author[4]{Arpit Bansal}
\author[1]{Zeyad Emam}
\author[2]{Furong Huang}
\author[2]{Micah Goldblum}
\author[2]{Tom Goldstein}

\affil[1]{Department of Mathematics, University of Maryland}
\affil[2]{Department of Computer Science, University of Maryland}
\affil[3]{Department of Robotics, University of Maryland}
\affil[4]{Department of Electrical and Computer Engineering, University of Maryland}

\makeatletter
\def\@maketitle{%
  \newpage
  \null
  \begin{center}%
  \let \footnote \thanks
    \hrule height 0.2em
      \vskip 1em%
    {\Large\bfseries\@title \par}%
    \vskip 0.5em
    \hrule height 0.1em
    \vskip 1.5em%
    {\large
      \lineskip .5em%
      \begin{tabular}[t]{c}%
        \@author
      \end{tabular}\par}%
    \vskip 1em%
  \end{center}%
  \par
  \vskip 1.5em}
\makeatother

\begin{document}

\maketitle

\section{Introduction}


In domains like computer vision, single and multi-agent games, and mathematical reasoning, classically trained models perform well on inputs from the same distribution used for training, but often fail to extrapolate their knowledge to more difficult tasks sampled from a different (but related) distribution. The goal of approaches like deep thinking and algorithm learning is to construct systems that achieve this extrapolation.   With this in mind, we detail several datasets intended to motivate and facilitate novel research into systems that generalize from easy training data to harder test examples. 

We present three datasets: \emph{Prefix Sums}, \emph{Mazes}, and \emph{Chess Puzzles}. We also provide an easy to install Python package and accompanying documentation to make training and testing on this data accessible.\footnote{The source code is available at \href{http://github.com/aks2203/easy-to-hard-data}{\texttt{http://github.com/aks2203/easy-to-hard-data}}.}

\section{Prefix Sums}
\label{sec:prefixsums}

The {\em Prefix Sums} dataset is meant to provide a simple toy baseline for testing new approaches, as models can be trained and tested rapidly on this dataset.   Inputs are lists of binary ($0/1$) numbers.  Each sample comes with a label/target, which is a list of the same length containing the cumulative sums modulo two for the input strings. For 52 different input lengths, we provide sets of 10,000 examples each. The shortest strings available are 16 bits, and we have every length through 64 bits as well as 72, 128, 256, and 512.

We refer to longer input strings as ``harder'' examples, which follows from the classical algorithms understanding of work required to compute the prefix sums. By offering so many different lengths, we provide many levels of difficulty.  

\subsection{Data Generation}

For each choice of the sample length $n,$ we produce 10,000 unique random strings, each containing $n$ binary numbers that represent a fair coin flip. We then compute their cumulative sums modulo two and save the input-output pairs. 

\subsection{Examples}

Examples from the sets with 16 bit and 28 bit inputs are shown below.

\begin{figure}[h]
    \centering
         \begin{tabular}{rl}
    Input: &  $[1, 0, 1, 0, 1, 0, 1, 1, 0, 0, 1, 1, 1, 0, 1, 1]$ \\
    Target: & $[1, 1, 0, 0, 1, 1, 0, 1, 1, 1, 0, 1, 0, 0, 1, 0]$\\
        Input: &  $[1, 0, 0, 1, 1, 0, 1, 1, 0, 1, 1, 0, 1, 0, 1, 1, 0, 0, 0, 1, 1, 0, 1, 0, 1, 1, 0, 0]$ \\
        Target: & $[1, 1, 1, 0, 1, 1, 0, 1, 1, 0, 1, 1, 0, 0, 1, 0, 0, 0, 0, 1, 0, 0, 1, 1, 0, 1, 1, 1]$
    \end{tabular}
    \caption{{\em Prefix Sums} input samples and their corresponding targets/labels. We provide multiple sets, each containing problems of a different length, and intend for users to train on shorter strings and test on longer ones. Examples from the sets of length 16 and 28 are shown above.}
    \label{fig:prefix_example18}
\end{figure}

\section{Mazes}
\label{sec:mazes}

The {\em Mazes} dataset contains images of mazes and their solutions. The solutions are represented as segmentation maps of the input pixels, with one class for pixels that are on the optimal (i.e., shortest) path from start to finish, and another class for those that are not on the optimal path. The inputs are three-channel (RGB) images.  The ``start'' of the maze is marked in red, the ``end'' of the maze is marked by a green square, and the walls of the maze are black.  Every square where the player is allowed to move is white.  Note that the ``start'' and ``end'' positions of an example can be swapped without changing the label; we do not care about the direction in which the path is walked. 

\subsection{Data Generation}

\begin{figure}
    \centering
    \includegraphics[width=0.9\textwidth]{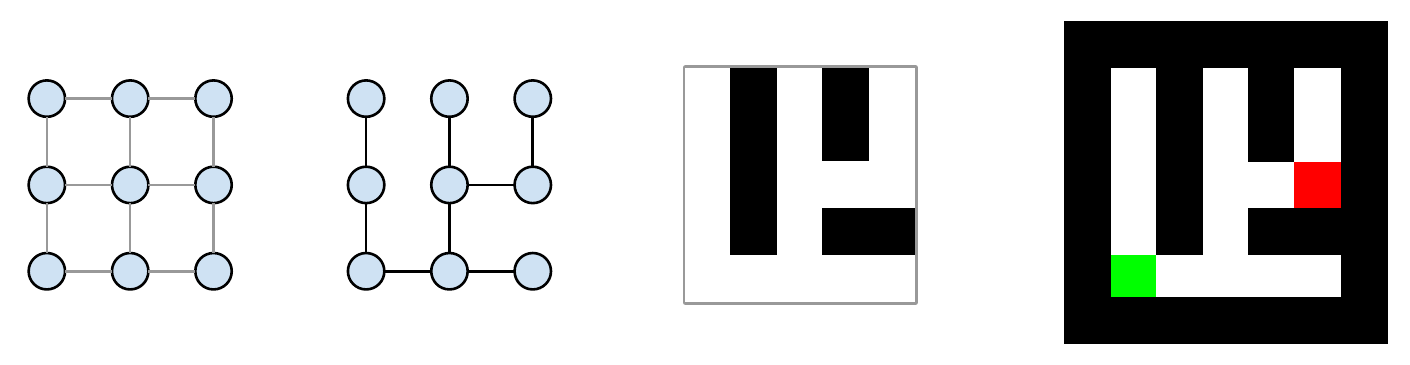}
    \caption{The \emph{Mazes} generation process for making a $5 \times 5$ maze. We start with a  $3 \times 3$ grid graph.  Each side of the grid graph contains 3 nodes and 2 edges ($3+2=5$ total elements).  We then produce a spanning tree for the graph using a randomized algorithm.  The tree encodes the allowed and forbidden paths.  This tree is then represented as a $5 \times 5$ array of maze cells.  Finally, we convert this to an image by representing each cell as a $2 \times 2$ array of pixels, and adding a 3-pixel border on each side. This creates an image representation that has $5\times 2 + 3 + 3 = 16$ pixels on each side. The green and red start and end cells are chosen at random.}
    \label{fig:maze_generation}
\end{figure}

We generate the mazes as abstract graphs, and then render them as images. We initialize a square grid graph, then use depth first search to find a subset of the edges that form a spanning tree. We randomly assign two nodes in the graph to be the end points of the maze. To render an image from this representation, we represent each node and each edge in the spanning tree with a white cell and each edge from the grid graph that is not in the spanning tree as a black cell. See Figure \ref{fig:maze_generation} for a depiction of each stage in the maze generation process. The code used was adapted from another maze generation repository \cite{hill2017making}. Finally, the solutions are generated with breadth-first search, using the image as input. By construction, the generated mazes admit a unique solution, in other words, there is exactly one path connecting the ``start'' and ``end'' positions. 

A note on the size of mazes: After creating an $n\times n$ maze, we represent the graph using $2\times 2=4$ pixels per cell, and add a 3 pixel wide black border on each side.  For this reason, a dataset of $n\times n$ mazes is represented using images of dimension $(2n+6)\times (2n+6)$.  For example, the $9\times 9$ \emph{Mazes} dataset contains images of size $24\times 24$ pixels.

\subsection{Examples}

\begin{figure}[ht]
\centering
\includegraphics[width=0.11\textwidth]{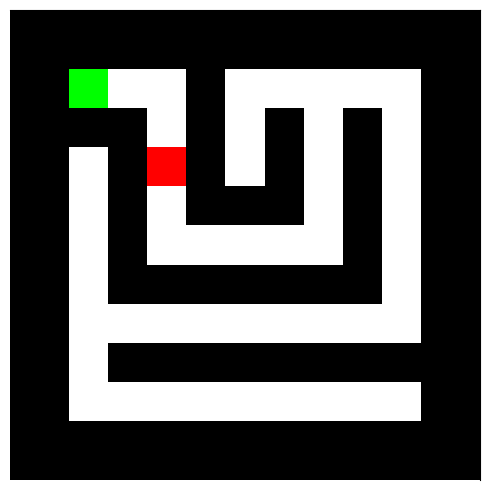}
\includegraphics[width=0.15\textwidth]{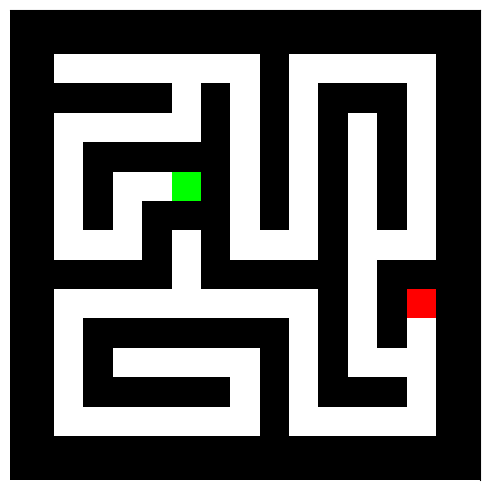}
\includegraphics[width=0.22\textwidth]{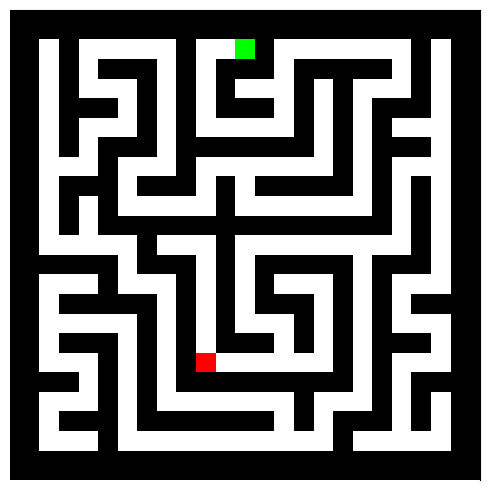} \\
\includegraphics[width=0.11\textwidth,valign=t]{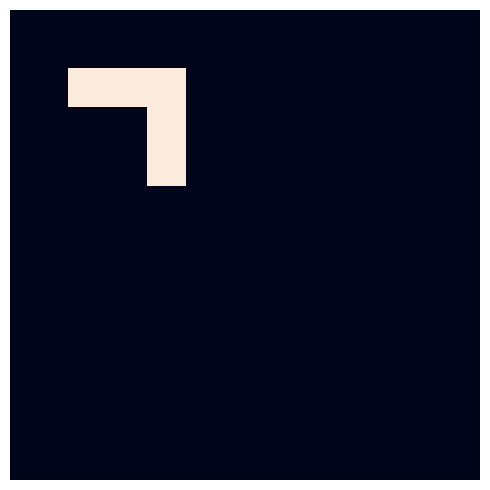}
\includegraphics[width=0.15\textwidth,valign=t]{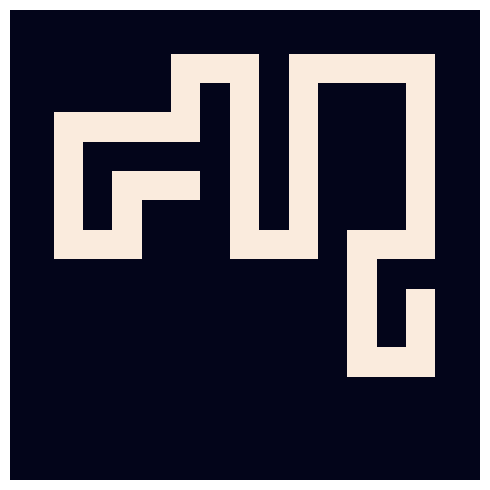}
\includegraphics[width=0.22\textwidth,valign=t]{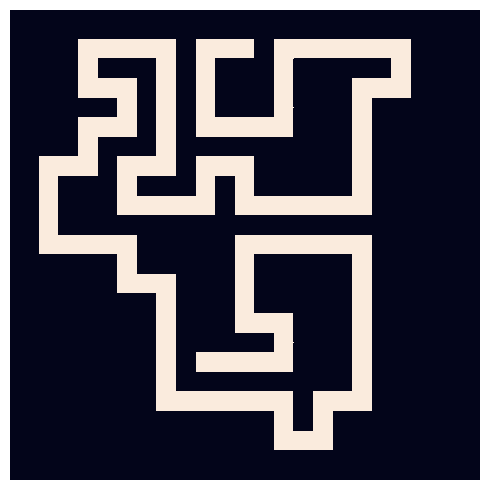}
\caption{Samples from \emph{Mazes}. Mazes (top) and their labels/solutions (bottom) of size $9 \times 9$, $13 \times 13$, and $21 \times 21$.}
\label{fig:mazes9}
\end{figure}

Figure \ref{fig:mazes9} shows examples of mazes and their solutions from datasets of three different difficulties.

\section{Chess Puzzles}
\label{sec:chess}

Training a full-scale chess game system is a complex task that requires large code bases and lots of compute resources. The chess puzzles compiled and provided by Lichess \cite{lichess} make for a much more accessible task. In these puzzles, the inputs are mid-game boards and the outputs are the unique best next move as determined by a state of the art chess engine. Here, we provide a \emph{Chess Puzzles} dataset that encodes the complex decision problems required to play chess, but is formatted as a simple pixel-wise classification problem that is easily used for training and testing. Our dataset consists of images representing game states (i.e., board configurations) and labels for each game state, which are 2D maps in which the source and destination positions for the optimal move are marked with a 1, and all other positions are marked with a 0.

\subsection{Data Generation}

By combing through 200,000,000 user games using the Stockfish 12/13 chess engines, Lichess creates puzzles consisting of sequences of unique best next moves. Once available for play on Lichess's website, a puzzle's initial Elo rating is then refined by having users with known ratings attempt to defeat the puzzle.\footnote{Elo is a system for rating players; one increases their Elo by besting other players with high Elo, and decreases their Elo by failing against lower rated players.  Chess puzzles are rated similarly by presenting them to players with known skill rating.} Interactions with hundreds/thousands of users eventually leave the puzzle with an equilibrated final rating, which quantifies difficulty. Each puzzle is tagged with the Forsyth-Edwards Notaion (FEN) representation of the board and an automatically generated short descriptor using code made available on the Lichess website. In other words, we can download a string that describes the current board, the best next move, and the difficulty rating for each of these puzzles. 

We use FEN information to create a Pytorch tensor that encodes the board state as a multi-channel ``image.''  This representation consists of twelve $8 \times 8$ channels, one for each class of white piece and one for each class of black piece. In particular, each channel has ones at the locations of the relevant pieces (e.g., pawns for one channel, rooks for another, etc..) and zeros elsewhere. The first six channels correspond to the player who is to move next. For this reason, a machine learning system that accepts this representation should not need to be explicitly told which player acts next, however a boolean flag for this purpose can be requested by the user when the dataset is constructed. Note, that our formulation of this dataset leaves out information about castling rights and \emph{En passant} -- a simplification of the game we make for convenience.

Each board state is labeled with a single-channel $8 \times 8$ tensor representing the solution to the problem. In these tensors there are zeros everywhere except on the origin and destination squares for the piece that should be moved. Again, note that the target itself does not possess the explicit information about what type of piece or which color is to move, but that information can be determined with the input and the identification of which color's turn it is.  

\subsection{Examples}

\begin{figure}[ht]
\centering
\includegraphics[width=0.32\textwidth]{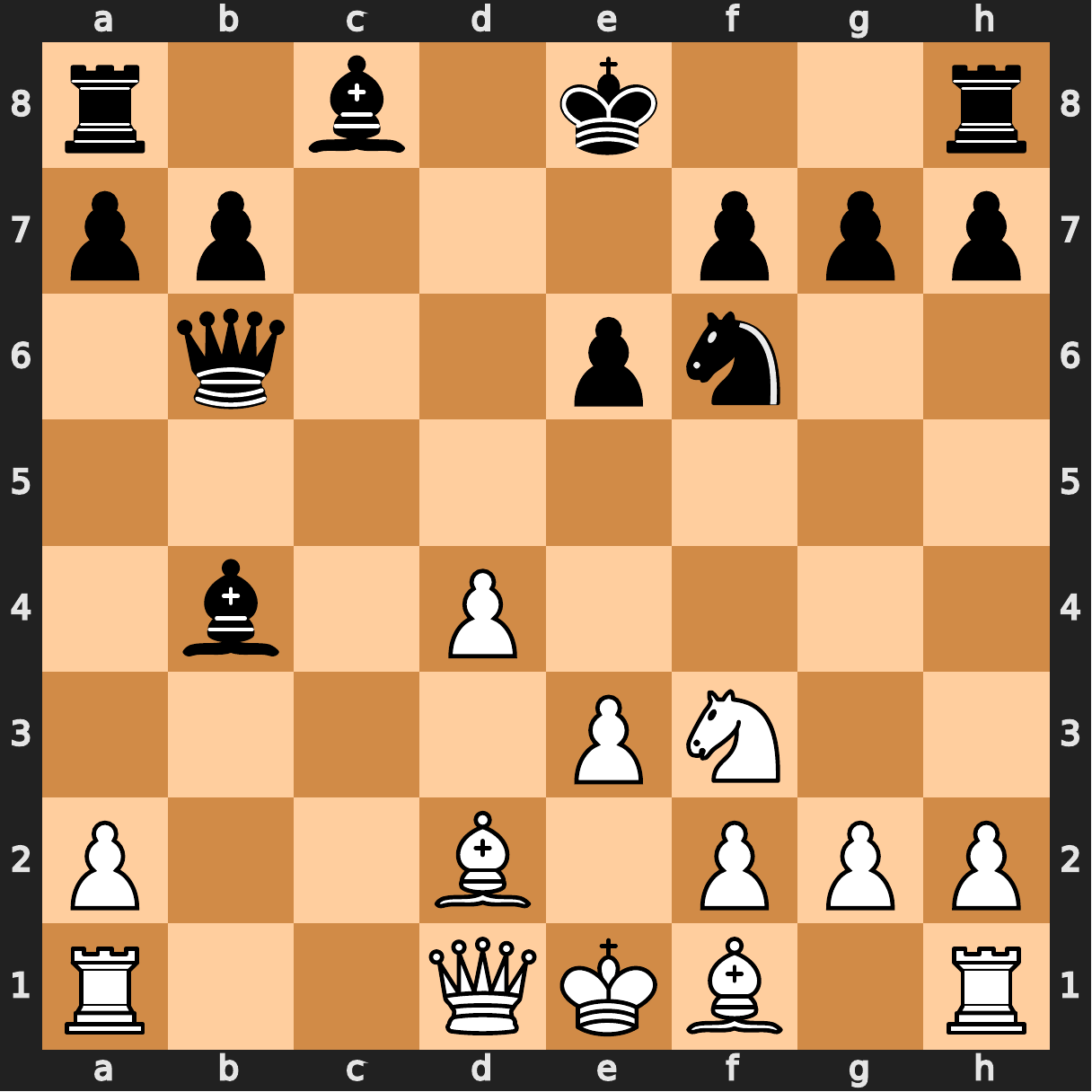}
\includegraphics[width=0.32\textwidth]{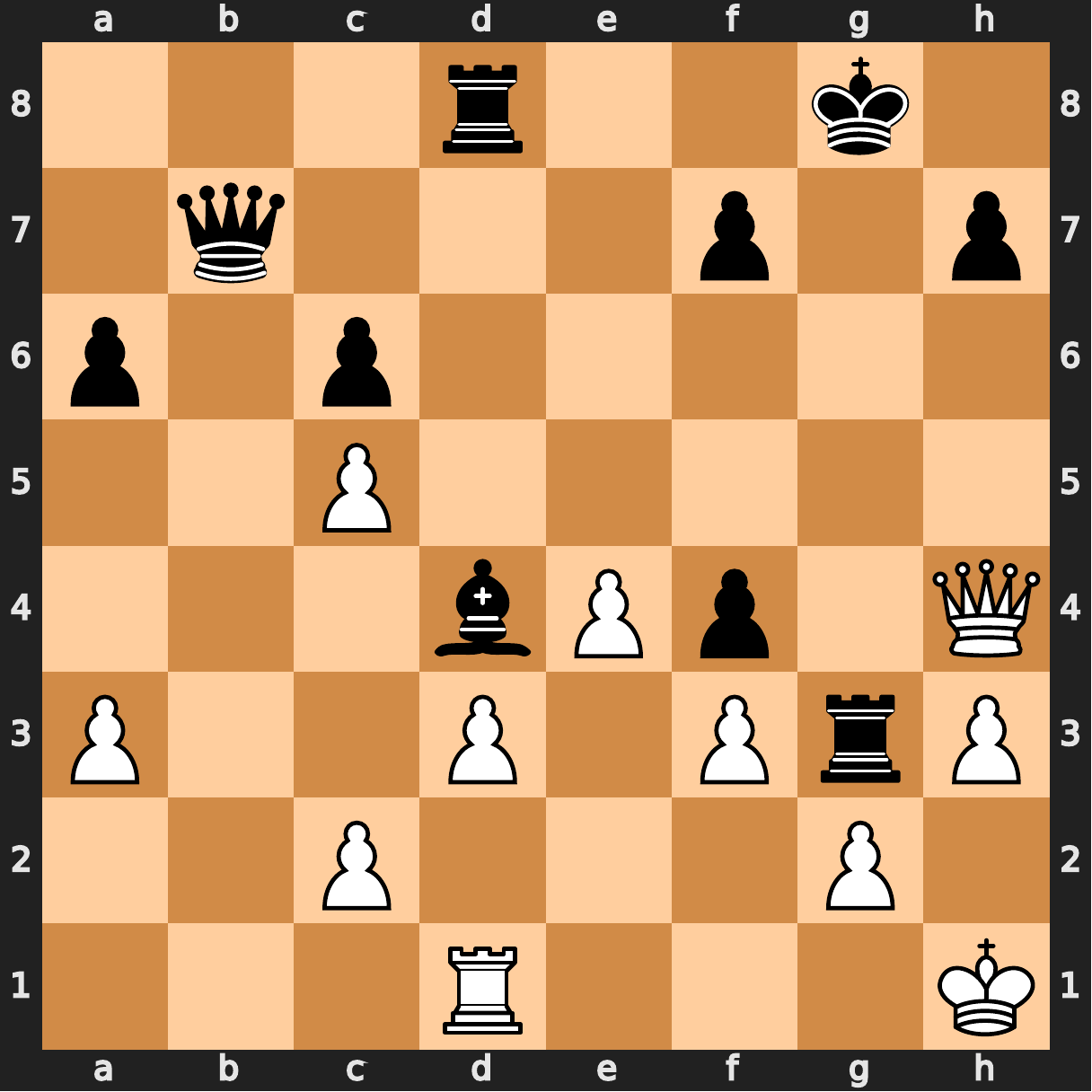}
\includegraphics[width=0.32\textwidth]{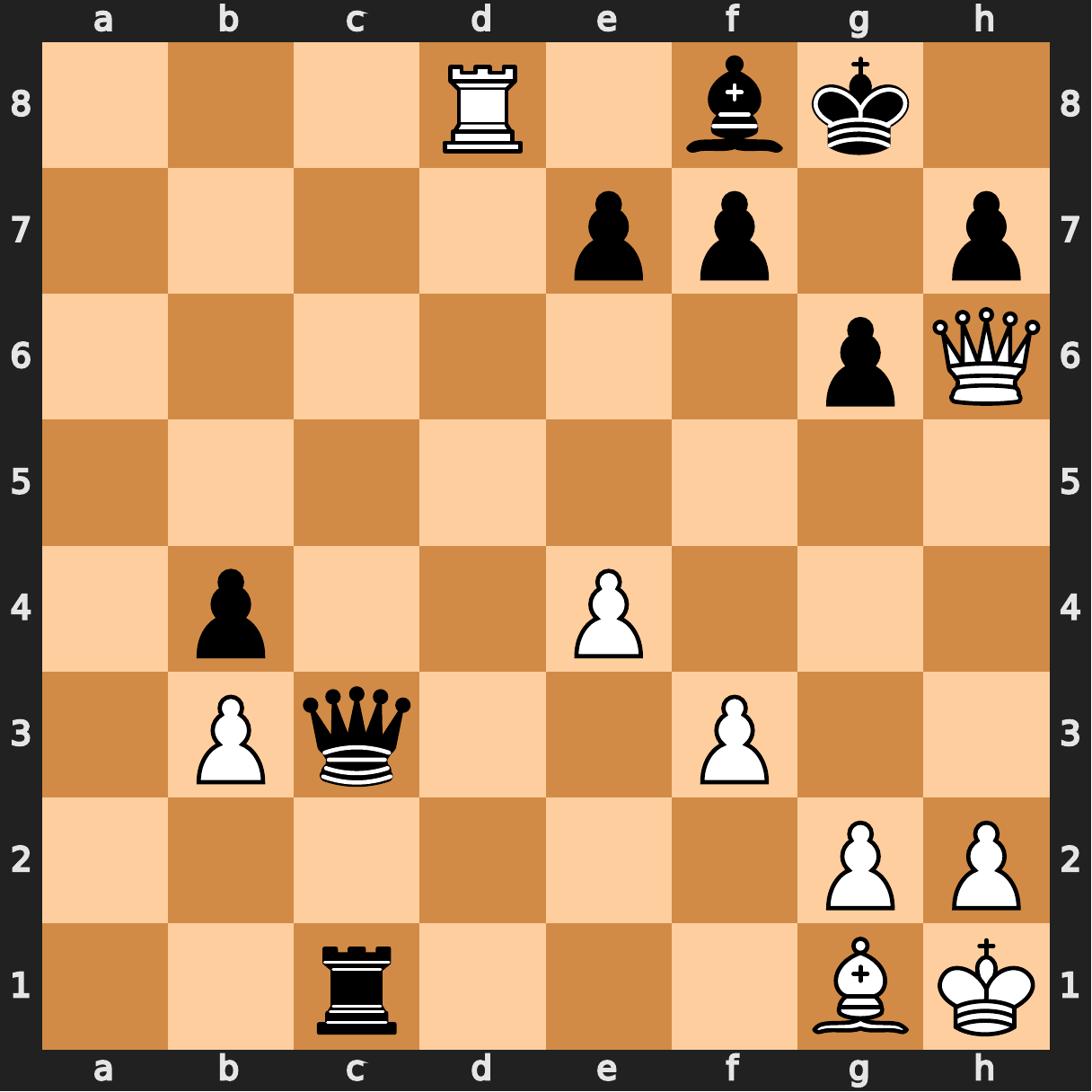}\\
\includegraphics[width=0.32\textwidth, trim=1.3cm 1.45cm 1.0cm 1cm, clip]{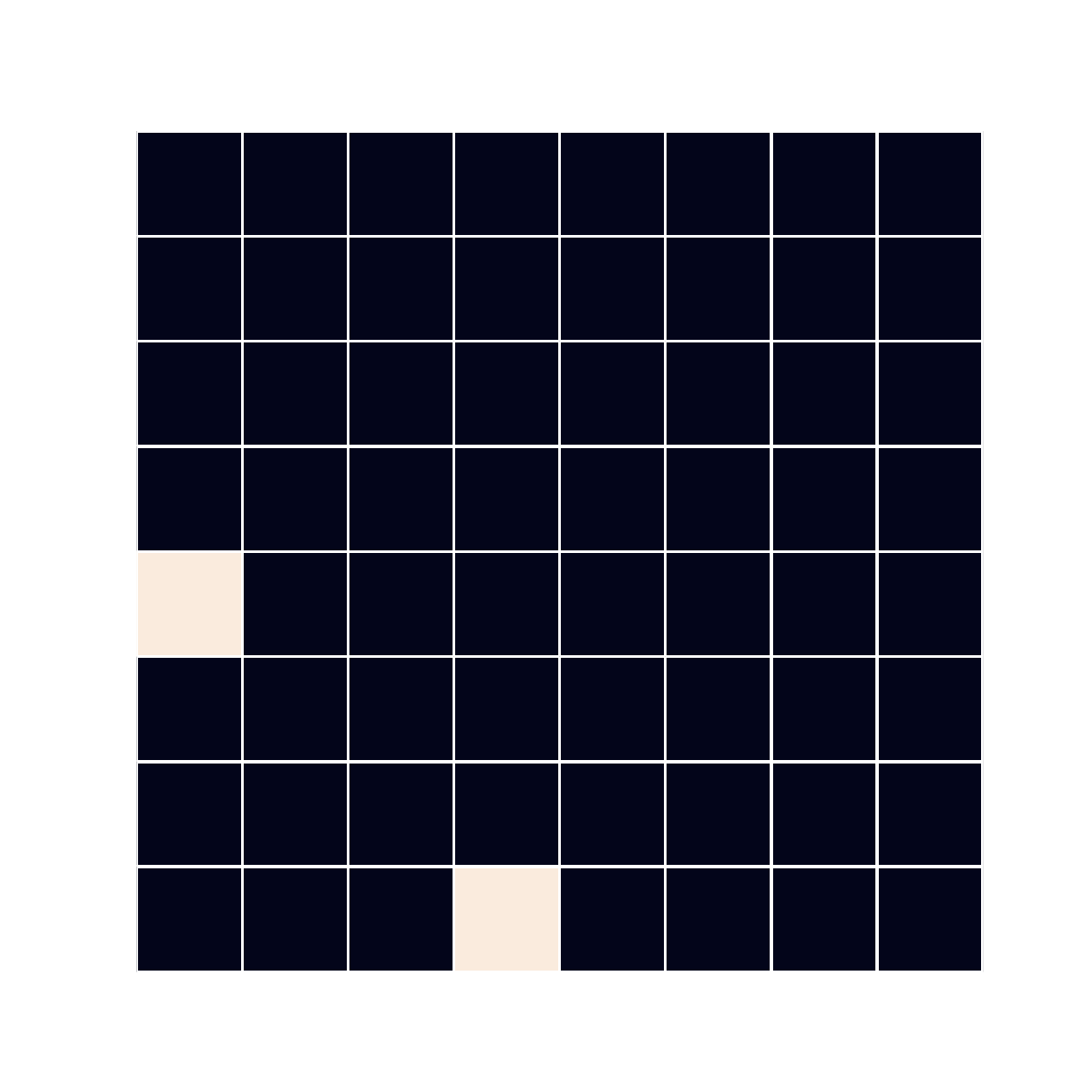}
\includegraphics[width=0.32\textwidth, trim=1.3cm 1.45cm 1cm 1cm, clip]{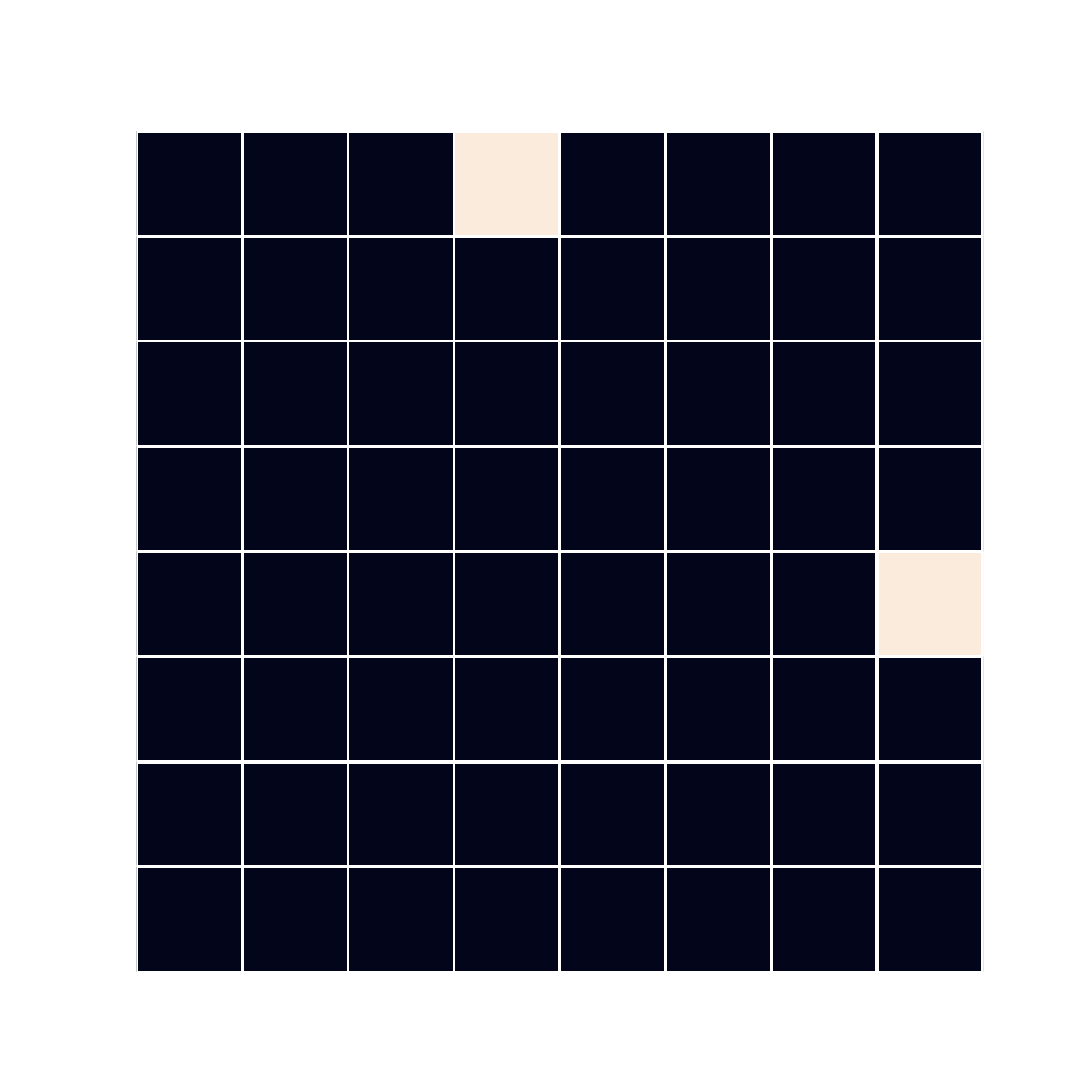}
\includegraphics[width=0.32\textwidth, trim=1.3cm 1.45cm 1cm 1cm, clip]{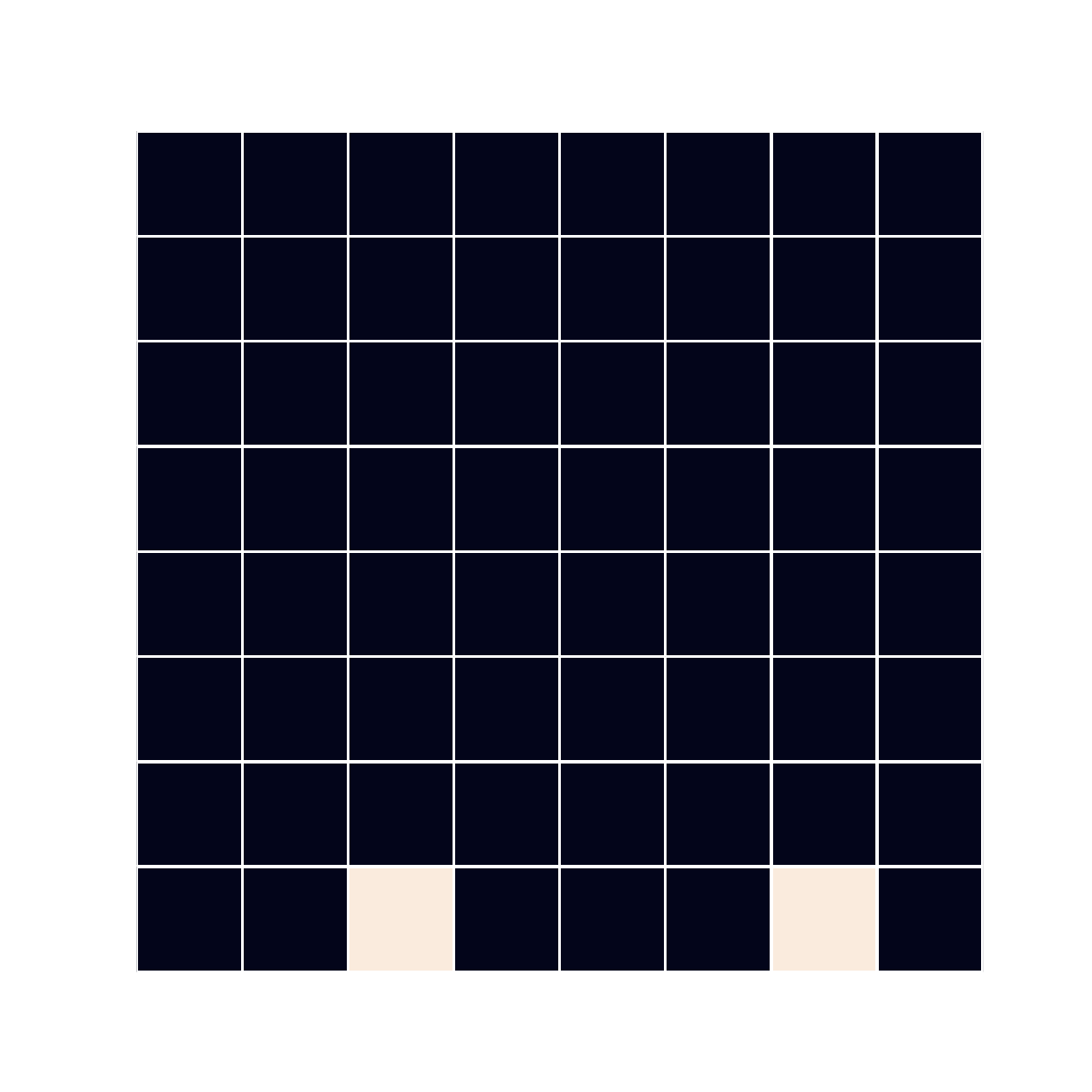}
\caption{Samples from \emph{Chess Puzzles}. ``Easy'' chess puzzles, each with their solution below.  The board state is represented as a $12\times 8\times 8$ tensor, where the first 6 maps encode the pieces belonging to the player to act next. The white player is acting in the two leftmost puzzles, while the black player acts in the puzzle on the right.  The solution is represented as a $1\times 8\times 8$ tensor that marks the start and end position of the piece to be moved.}
\label{fig:chess0}
\end{figure}

\section{Python Package}
\label{sec:python}

The Python package for these datasets is available through the Package Installer for Python (pip). We provide Python classes for each problem type, and the class constructors accept an argument specifying the desired difficulty for the dataset. The documentation and code is available \href{http://github.com/aks2203/easy-to-hard-data}{\texttt{here}}.\footnote{\texttt{https://pypi.org/project/easy-to-hard-data/}}

When a dataset object is created, the user can point the constructor to the location of the downloaded raw data, or else a flag can be set to perform this download automatically.  

In addition to initializing dataset objects, the repository also houses code for generating the data and code to make visualizations for mazes and chess. The dataset objects rely on standard Pytorch features, and should be easy to modify to create variations on the standard benchmark problems.

\bibliographystyle{abbrv}
\bibliography{main}

\end{document}